\pdfoutput=1

\documentclass[11pt]{article}

\usepackage{emnlp2021}

\usepackage{times}
\usepackage{latexsym}

\usepackage[T1]{fontenc}

\usepackage[utf8]{inputenc}

\usepackage{microtype}

\usepackage{CJK}
\usepackage{multirow}
\usepackage{soul}
\usepackage{graphicx}
\usepackage{microtype}
\usepackage{graphicx}
\usepackage{amsmath}
\usepackage{amsfonts}
\usepackage{amssymb}
\usepackage{textcomp}
\usepackage{xcolor}
\usepackage{cleveref}
\usepackage{mathrsfs}
\usepackage{amsthm}
\usepackage{bm}
\usepackage{booktabs}
\usepackage{enumitem}
\usepackage{times}
\usepackage{xspace}

\crefformat{section}{\S#2#1#3} 
\crefformat{subsection}{\S#2#1#3}
\crefformat{subsubsection}{\S#2#1#3}
\usepackage[textwidth=0.8in,textsize=scriptsize]{todonotes}

\newcommand{\ourtask}{Narrative Incoherence Detection\xspace}

\begin{document}

\begin{CJK*}{UTF8}{}
\title{Narrative Incoherence Detection}

\CJKfamily{gbsn}
\author{
Deng Cai$^\flat$\quad Yizhe Zhang$^\natural$\quad Yichen Huang (黄溢辰)$^\sharp$\quad Wai Lam$^\flat$\quad Bill Dolan$^\natural$ \\ 
$^\flat$The Chinese University of Hong Kong \\
$^\natural$Microsoft Research, Redmond \\
$^\sharp$Center for Theoretical Physics, Massachusetts Institute of Technology\\
{\tt \{dcai,wlam\}@se.cuhk.edu.hk}\\
{\tt \{yizhe.zhang,billdol\}@microsoft.com}\\
{\tt yichuang@mit.edu}}

\maketitle
\end{CJK*}
	\begin{abstract}
		We propose the task of narrative incoherence detection as a new arena for inter-sentential semantic understanding: Given a multi-sentence narrative, decide whether there exist any semantic discrepancies in the narrative flow. Specifically, we focus on the missing sentence and discordant sentence detection. Despite its simple setup, this task is challenging as the model needs to understand and analyze a multi-sentence narrative, and predict incoherence at the sentence level. As an initial step towards this task, we implement several baselines either directly analyzing the raw text (\textit{token-level}) or analyzing learned sentence representations (\textit{sentence-level}). We observe that while token-level modeling has better performance when the input contains fewer sentences, sentence-level modeling performs better on longer narratives and possesses an advantage in efficiency and flexibility. Pre-training on large-scale data and auxiliary sentence prediction training objective further boost the detection performance of the sentence-level model.
	\end{abstract}
    \section{Introduction}
    \begin{figure}[t]
		\centering
		\includegraphics[width=.85\linewidth]{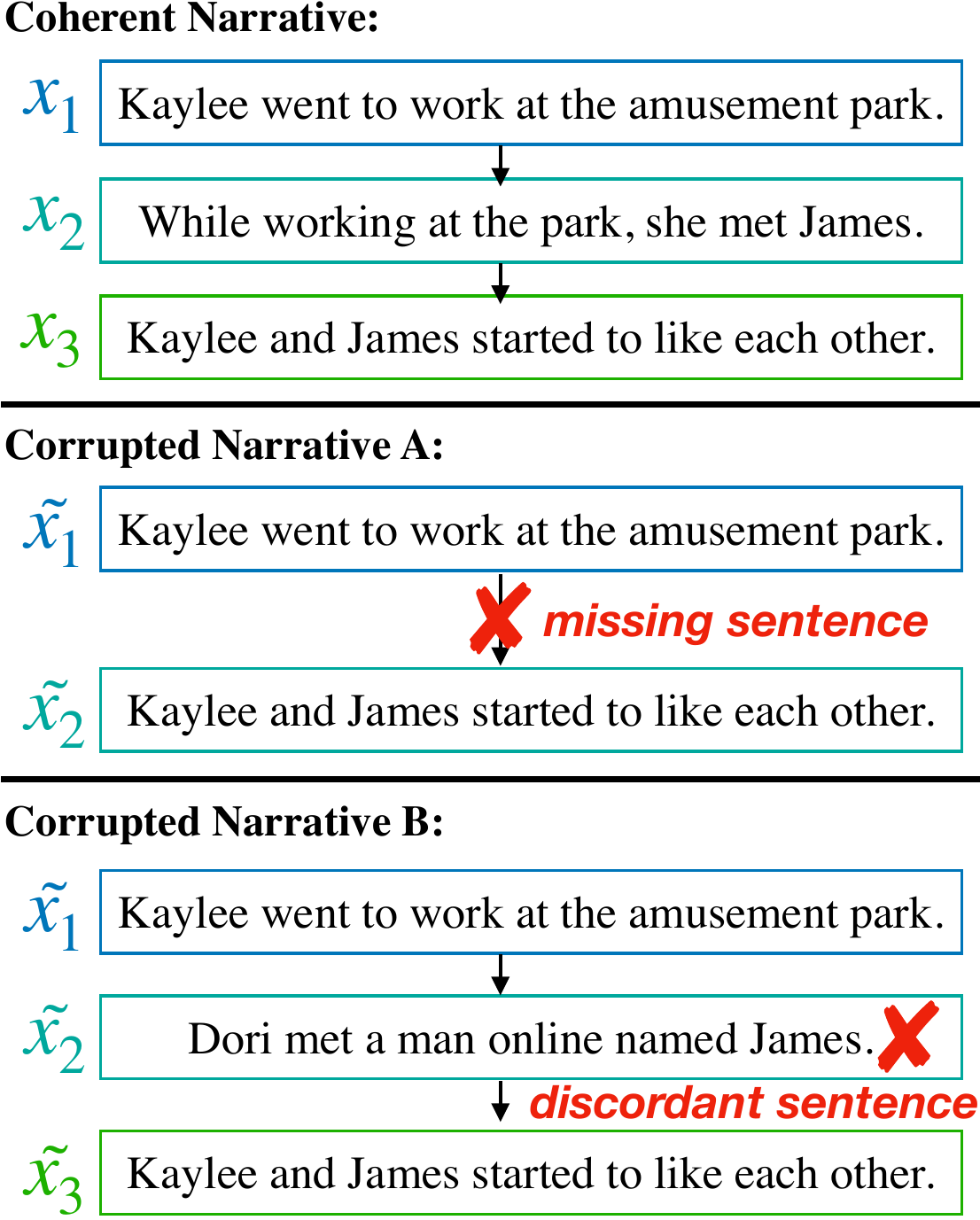}
		\caption{Illustration of our Narrative Incoherence Detection tasks. Note that a narrative contains more sentences in real cases.}
		\label{example}
 		\vspace{-4mm}
	\end{figure}
    
	Long-form text understanding and generation are of great interest yet remain key challenges in natural language processing. Recent years have witnessed significant advances in natural language understanding and generation thanks to large-scale pre-trained language models, such as \textsc{Bert} \cite{Devlin2019BERTPO}, \textsc{Gpt-2} \cite{Radford2019LanguageMA}, and \textsc{Gpt-3} \cite{Brown2020LanguageMA}. These models can produce individual sentences that are grammatical and fluent \cite{donahue-etal-2020-enabling}. However, they are not designed to capture the inter-sentential semantic flow among sentences and may have difficulty in analyzing or producing a coherent multi-sentence narrative. For example, it has been reported that when generating long text, models such as \textsc{Gpt-2} may easily get stuck in repetitions \cite{Holtzman2020TheCC}.

	The aforementioned problem can be partially attributed to the lack of specific machinery to extract and characterize the inter-sentential semantic flow in multi-sentence text \cite{ippolito-etal-2020-toward, Kang2020PlanAS}. Despite its importance, learning inter-sentential coherence remains an open challenge, as it requires $(i)$ understanding, extracting, and representing the high-level semantic flow for a given text; $(ii)$ at the discourse level, addressing logical, commonsense reasoning, and planning. In addition, evaluating the discourse-level understanding capability of a given model is also an open problem. Previous work has tackled specific aspects of this challenge including understanding tasks such as causal reasoning \cite{kang-etal-2017-detecting}, abductive reasoning \cite{Bhagavatula2020AbductiveCR}, sentence ordering \cite{Barzilay2008ModelingLC}, narrative cloze test \cite{Chambers2008UnsupervisedLO,Mostafazadeh2016ACA}, reading comprehension \cite{wang2019superglue}, and generation tasks such as story generation \cite{Fan2019StrategiesFS}, text infilling \cite{Hua2019SentenceLevelCP,Huang2020INSETSI,kang-hovy-2019-linguistic}, and counterfactual plot rewriting \cite{Qin2019CounterfactualSR}.
	
	In this paper, we propose a new task called \textit{\ourtask}~as a new testbed for benchmarking a model's ability to capture the inter-sentential semantic flow of narratives. As illustrated in Figure~\ref{example}, we consider a multi-sentence passage where a certain amount of semantic discrepancies are caused by missing some sentences or containing some discordant sentences. The task is to identify the positions of the missing/discordant sentences that introduce the semantic discrepancies. Compared with existing tasks, our task enjoys the following merits: $(i)$ It is conceptually simple, and the training and testing datasets can be created with much less human annotation effort. $(ii)$ It can cover a broad range of inter-sentential understanding challenges (\textit{e.g.}, logical, commonsense, causal, and temporal reasoning), which are associated with the semantic coherence in different narratives. $(iii)$ The evaluation is straightforward and objective. The performances of predictive models for our task can be measured and compared using a set of classification metrics. Furthermore, the proposed task has independent merits in practice. The models developed for the task can potentially increase the functionality of intelligent editing assistants. For example, they can be used for document proofreading by $(i)$ detecting missing sentences needed to bridge discontinuous context, and $(ii)$ detecting problematic sentences that compromise the coherence of a narrative, especially when a document is written by multiple authors.
	
	As an initial step towards solving the proposed task, we investigate two types of baseline approaches. These include two popular modeling paradigms in the current literature: \textit{token-level} and \textit{sentence-level} approaches. The token-level approach directly concatenates the input sentences in an orderly manner and processes them as a flat sequence. It fine-tunes \textsc{Bert} \cite{Devlin2019BERTPO} with necessary modifications to the input format to accommodate our task. In contrast, the sentence-level approach uses sentences as atomic units. It views a multi-sentence narrative as a sequence of pre-trained sentence embeddings and processes them in a Transformer \cite{vaswani2017attention} that operates at the sentence level. The sentence-level approach is more computationally efficient than the token-level approach especially when the input narrative is long. To take advantage of the efficiency, we pre-train the sentence-level models on massive data and observe significant performance improvements. Furthermore, the sentence-level approach opens up the possibility of joint incoherence detection and sentence prediction learning (i.e, learning to predict appropriate sentences to (re)fill corrupted positions) with little extra architectural design and computational overhead. Our experiments show that a joint model performs adequately well in both tasks.
	
	Our contributions are summarized as follows:
	\begin{itemize}[wide=0\parindent,noitemsep,topsep=0em]
	    \item We introduce \textit{\ourtask} as a new task for inter-sentential semantic understanding. We collect four medium-size test sets using crowd-sourcing, covering different incoherence types and text lengths. These test sets can be used for future model development and evaluation.
	    \item We establish token-level and sentence-level baselines and compare them in extensive experiments. We show that the token-level approach has better detection accuracy with shorter input while the sentence-level approach is more accurate for longer input and more efficient. We observe that pre-training sentence-level models with large external corpus improves the performances.
	    \item We show that the sentence-level baselines can be further enhanced by exploiting the synergy between \ourtask and the corresponding sentence prediction task.
	\end{itemize}
	\section{Problem Statement}
	The task of \ourtask~ is to take as input a multi-sentence text $x$ containing one or more semantic discrepancies, and return the positions of the discrepancies. To focus on discourse-level coherence, we assume that each sentences $x_k$ in the input prose $x=(x_1, x_2, \ldots, x_N)$ is grammatical and fluent. Each sentence $x_k$ itself is a sequence of tokens $x_k = (x_{k}^1, x_k^2, \ldots, x_{k}^{L_{k}})$ of length $L_k$. We consider two scenarios: \textit{missing sentence detection} and \textit{discordant sentence detection}. These two scenarios correspond to the ``insertion'' and ``replacement'' needs for text editing.
	\paragraph{Missing Sentence Detection (MSD)}
	The first case we consider is MSD, where some semantic gaps are caused by missing bridging sentences (Figure~\ref{example}, middle). That is, the input paragraph $x=(x_1, x_2, \ldots, x_N)$ is a sub-sequence of a complete and coherent paragraph $\tilde{x} = (\tilde{x}_1, \tilde{x}_2, \ldots, \tilde{x}_M)$, where $M-N$ intermediate sentences are missing. Formally, we have $x_i = \tilde{x} _{\phi_i}$, and $1=\phi_1 < \phi_2 < \cdots < \phi_{N}=M$ ($\phi_{1:N}$ is a strictly increasing sequence of indices). Taking $x$ as input, the goal is to predict the label $y_k \in \{0, 1\}$, indicating whether there is a missing sentence (semantic gap) between $x_k (\tilde{x}_{\phi_k})$ and $x_{k+1} (\tilde{x}_{\phi_{k+1}})$, for all $k \in [1, N)$.
	\paragraph{Discordant Sentence Detection (DSD)}
	Our second scenario is DSD, where some sentences in a paragraph are discordant to the context (Figure~\ref{example}, bottom). Specifically, the input paragraph $x$ is assumed to be a corrupted version of a normal and coherent paragraph $\tilde{x}$, where one or more sentences are replaced with confounding sentences. To focus on inter-sentential semantic discrepancy rather than lexical or grammatical issues, the confounding sentences are grammatical and fluent but semantically incongruous with the context. The goal is to predict the label $y_k \in \{0, 1\}$, indicating whether $x_k$ is such a problematic sentence, for all $k \in [1, N]$.
	\begin{figure*}[t]
		\centering
		\includegraphics[width=0.90\linewidth]{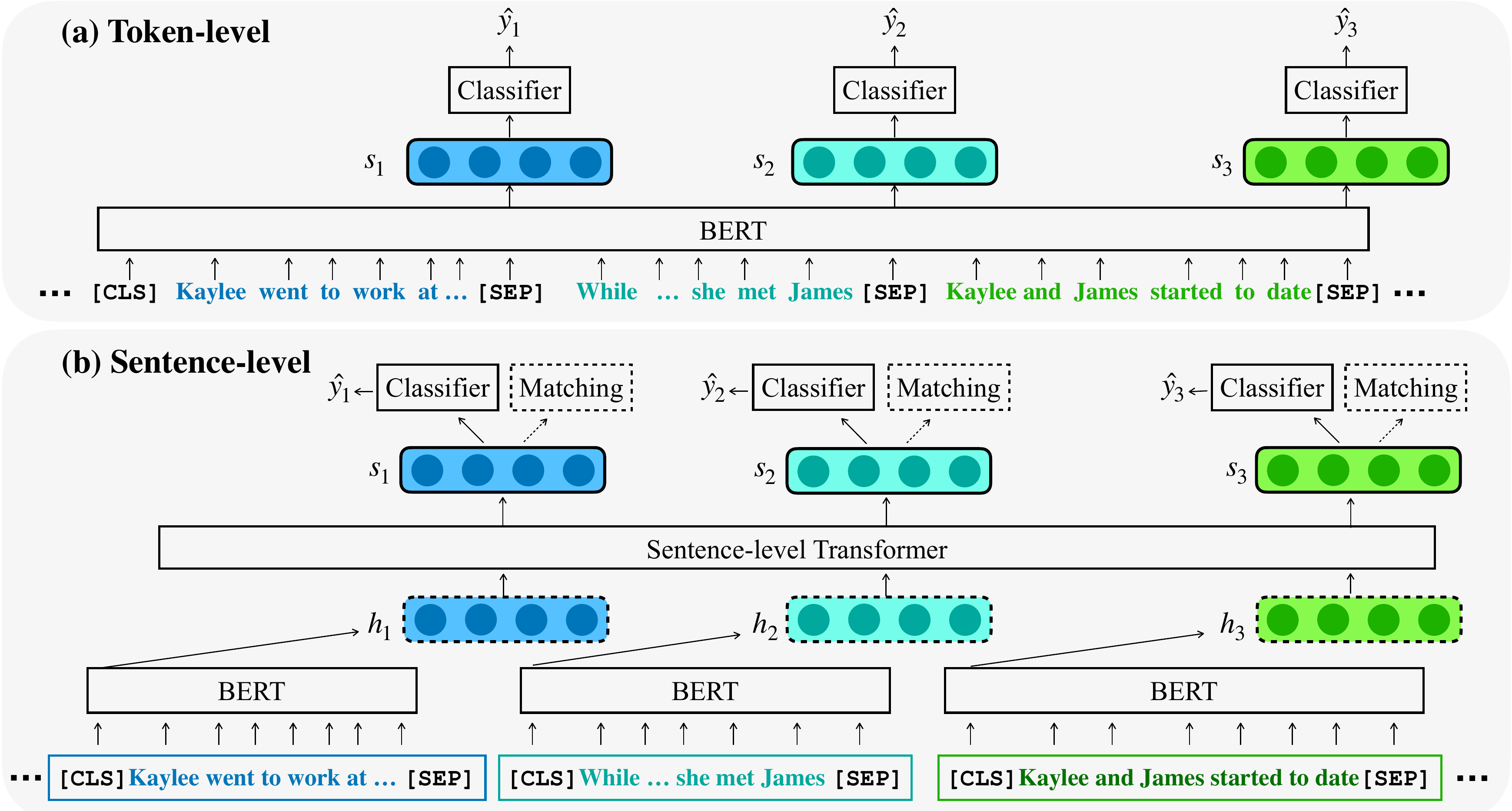}
		\caption{The architectures of the token-level model (a) and sentence-level model (b).}
		\label{arch}
	\end{figure*}
	\section{Datasets}
			\begin{table}[t]
		\centering
		\small
		\begin{tabular}{c|c|c}
			\hline
			Dataset & \textsc{TimeTravel} & \textsc{TripAdvisor} \\
			\hline
			\# Training Paragraph & 126,524 & 140,910\\
			\# Dev Paragraph & 7,484  & 17,613\\
			\# Test Paragraph & 7,484\ & 17,613\\
			\hline
		\end{tabular}
		\caption{Dataset statistics}
		\label{datasets}
		\vspace{-3mm}
	\end{table}
		\begin{table}[t]
		\centering
		\small
		\begin{tabular}{c|c|c}
			\hline
			Task & \textsc{TimeTravel} & \textsc{TripAdvisor} \\
			\hline
			MSD & 1,200 & 500\\
			DSD & 2,000 & 500 \\
			\hline
		\end{tabular}
		\caption{The sizes of our test sets.}
		\vspace{-4mm}
		\label{testsets}
	\end{table}
    \label{data-setup}
	\paragraph{Data Preparation} We create benchmark datasets for our study from two existing datasets: \textsc{TimeTravel} \cite{Qin2019CounterfactualSR} and \textsc{TripAdvisor} \cite{Wang2010LatentAR}. The \textsc{TimeTravel} dataset is an expansion of the \textsc{ROC} dataset \cite{Mostafazadeh2016ACA}. It contains five-sentence self-contained stories. The \textsc{TripAdvisor} dataset is a collection of hotel reviews and we use pre-processed data from \newcite{Huang2020INSETSI}. Both \textsc{TimeTravel} and \textsc{TripAdvisor} consist of compact narratives that follow certain logic flows. The original corpus statistics are shown in Table \ref{datasets}. Based on these narratives, we create data instances for our task using the following procedures.
	
	First, we randomly sample a consecutive segment of $M$ sentences from each prose in the original corpus. We then randomly pick $K$ sentences out of the $M$ sentences to create the altered text. For MSD, the $K$ selected sentences are removed and the remaining sentences are concatenated.\footnote{To reduce ambiguity, in all MSD instances, we do not remove the first or last sentence in the original segment. Also, when more than one sentence is removed, the removed sentences are not adjacent.} For DSD, we replace each of the $K$ selected sentences with a confounding sentence. The confounding sentence of a given sentence $x$ is obtained by the following search: (1) we first select the top $100$ most similar sentences from the entire corpus using the fast BM25 retrieval~\cite{robertson2009probabilistic}. (2) we then choose the highest-ranked sentence $\tilde{x}$ according to BM25, under the constraint that $sim(x, \tilde{x})<\tau$ where the similarity $sim$ is measured by \textsc{BertScore}~\cite{zhang2019bertscore} and $\tau$ is empirically chosen to be $0.7$.
	
	For the \textsc{TimeTravel} dataset, we set $N=5$ and $M=1$, resulting in a missing/discordant sentence rate of $1/5=20\%$. For the \textsc{TripAdvisor} dataset, we set $N=8$ and $M=2$ so that the missing/discordant sentence rate of $2/8=25\%$. Note that the missing/discordant sentence rate should not be set too high. This is because when most sentences are removed or replaced by confounding ones, the original narrative will be completely broken and the detection task would be extremely ambiguous or even impossible.
	
	\paragraph{Human Annotation} There are two practical issues regarding the data preparation described above: (1) The altered text might still be a coherent narrative in some cases. For example, when we remove a particular sentence from a normal paragraph, some missing sentences may introduce semantic gap, while some may not affect the coherence. (2) The positions of missing/discordant sentences can sometimes be ambiguous. For example, in DSD, when only two sentences contradict each other while both being compatible to other sentences, it is reasonable to label either one of them as discordant. Despite the noises in the automatically created data, from a statistical point of view, the distribution-wise difference between the original and altered texts may still provides a salient signal for learning and testing. Therefore, we use the automatically created data for training and development.
	
	However, we want our test sets to be less noisy and of high quality. To this end, we recruit crowd-workers on Amazon Mechanical Turk for annotation verification. We prepare a set of candidate test instances from the original test sets, and present each of them to four expert judges who passed a screening task. Each instance is about the incoherence judgement on a specific slot/sentence position. To ensure quality, a candidate test instance is included in our final test sets if and only if at least three of the four judges agree with the ``ground truth'' label in automatic data creation. The sizes of final test sets are shown in Table \ref{testsets}.
	
	We also establish human baselines by presenting each instance in our final test sets to another three expert workers and report the average performance (in Table \ref{results-msd-dsd}). Full details about human annotation are provided in the Appendix.
	\section{Baseline Methods}
	We consider two modeling paradigms as baseline approaches to tackle the \ourtask~task. First, we develop a \textit{token-level} approach where the input text is directly processed as a sequence of individual tokens. Second, we explore several strategies of \textit{sentence-level} approaches: The input text is first mapped to a sequence of sentence representations, and then a sentence-level model operates on the sentence representations for incoherence detection.
	
	\subsection{Token-Level Approach}
	The token-level model takes the sequence of tokens $x$ as input. Suppose that the input text $x$ has $N$ sentences in total and the length of the $k$-th sentence is $L_{k}$. The total number of tokens is $\sum_{k=1}^N L_{k}$. We train our model by fine-tuning the pre-trained \textsc{Bert}. Figure \ref{arch}(a) illustrates the architecture of the token-level approach.
	
	\ourtask~can be regarded as a sentence-level tagging task. However, \textsc{Bert} is pre-trained as a masked language model, and the output vectors are aligned with tokens rather than sentences. To represent the slots between sentences for MSD (or individual sentences for DSD), we insert $N-1$ indicator symbols (\texttt{[SEP]}) between adjacent sentences. In addition, one \texttt{[CLS]} and one \texttt{[SEP]} are added to the beginning and end of each sequence, respectively. The resulting input token sequence is fed into \textsc{Bert} to produce the contextualized vector representations for each token. The resultant vector $s_k$ corresponding to the $k$-th \texttt{[SEP]} symbol is used as the representation of the slot between $x_k$ and $x_{k+1}$ for MSD, or the sentence representation of $x_k$ for DSD.
	
	The vectors $\{s_k\}_{k=1}^N$ are then passed through a MLP layer ($\texttt{MLP}_d$) and a sigmoid layer to generate the normalized scores to predict the binary labels $\hat{y} = \{y_1,\ldots,y_N\}$:
	\begin{equation}
	    p(y_k=1) = \sigma (\texttt{MLP}_d(s_k)).
	    \label{eq-detect}
	\end{equation}
	The cross-entropy between the hypothesis $\hat{y}$ and the ground truth $y$ is used as the training objective.
	
	Note that for all predictions, the model has access to \textit{bi-directional} context. However, each prediction is conditionally independent given the contextualized output vectors.
	\subsection{Sentence-Level Approach}
	Despite its simplicity, the token-level approach is costly with long-form text, as computation and memory scale quadratically with the number of tokens in a vanilla transformer. As an alternative, we also develop sentence-level models as additional baselines, which treat the input as a sequence of sentence embeddings. The detection then directly operates at the sentence level by considering each sentence as an atomic unit. This is more efficient than the token-level approach, as the sequence length becomes the number of sentences $N$.
	
	\paragraph{Model Architecture}
	Specifically, we take advantage of pre-computed sentence representations from existing representation learning models. In our experiments, we use the \texttt{[CLS]} embedding of the last layer of a pre-trained bidirectional language model, \textsc{Bert} \cite{Devlin2019BERTPO}. The architecture of our model is shown in Figure \ref{arch}(b). The sentences in the input paragraph $(x_1, x_2, \ldots, x_N)$ are first mapped into vector representations $(h_1, h_2, \ldots, h_N)$ independently. Then, we use a sentence-level Transformer to exploit the associations among sentence embeddings.

	In a similar manner to the token-level approach (Eq. \eqref{eq-detect}), the final output vector $s_k$ corresponding to the $k$-th sentence is fed into a binary classifier in order to predict whether there is missing text between $x_k$ and $x_{k+1}$ in MSD, or whether $x_k$ is a discordant sentence in DSD.
	
	Sentence-level models have three unique features compared with the token-level approach. $(i)$ They concentrate on inter-sentence coherence and long-term semantic flow rather than word-level fluency and co-occurrence; $(ii)$ They take as input fixed sentence embeddings, which can be pre-computed, saved, and reused;\footnote{One may think of fine-tuning the sentence embedding with the detection objective. However, this reduces the efficiency advantage of the sentence-level approach.} $(iii)$ They decouple sentence representation learning and discourse-level understanding, and thus the two components can be separately optimized with additional goals. 
	
	These features of sentence-level modeling allow two extensions as detailed in the following: \textit{Pre-training at Scale} and \textit{Semantic Matching as an Auxiliary Task}.
	
	\paragraph{Pre-training at Scale} Sentence-level modeling requires much less computation and memory compared with token-level modeling. Specifically, suppose that a document contains $N$ sentences, each of which has $L$ tokens. The time complexity is reduced from $O(N^2L^2)$ (token-level) to $O(N^2 + NL^2)$. Moreover, sentence embeddings can be pre-computed and stored for later use. The time complexity for the remaining sentence-level transformer is only $O(N^2)$. This is particularly useful in practice as we train models on the same dataset for many epochs or in various setups.
	
	Due to the relatively cheap cost of sentence-level modeling, we pre-train our sentence-level model on large corpora such as \textsc{Stories} \cite{Trinh2018ASM} as detailed in \cref{exp-setup}. Note that such a pre-training step is prohibitively expensive for the token-level approach.
	
	\paragraph{Semantic Matching as an Auxiliary Task} A follow-up task to the \ourtask~task is \textit{sentence prediction} \cite{Huang2020INSETSI}, where the model generates a missing intermediate sentence (MSD) or a substitute of the current sentence (DSD) given the position of the missing/discordant sentence. The generation task and our detection tasks are highly relevant and partially entail each other. For example, predicting whether a position requires an additional bridging sentence might requiring knowing, to some extent, what information is missing and what needs to be interpolated to complete the semantic flow. To examine and potentially exploit the natural synergy between these two tasks, we formulate the task of \textit{sentence prediction} as an auxiliary semantic matching (SM) objective in our sentence-level framework with minimal architectural changes.
	
	Specifically, semantic matching can be performed by applying an additional MLP layer ($\texttt{MLP}_{sm}$) to the slot/sentence representation $s_{k}$:
	\begin{equation}
	    \hat h_k = \texttt{MLP}_{sm}(s_k).
	    \label{eq-generate}
	\end{equation}
	The auxiliary semantic matching objective is defined as the cosine distance between the prediction $\hat h_k$ and the embedding of the ground truth sentence. Note that the semantic matching module reuses the same hidden representations $s_k$ from the sentence-level transformer and runs in parallel with the detection classifier. The additional computational overhead of the SM objective is negligible.
	
    \subsection{Text Generation as side task}
    \label{text-gen}
    Regardless of improving \ourtask, the above auxiliary semantic matching objective further motivates us to explore the possibility of joint \ourtask~and text generation task. These two tasks are organic counterpart to each other and evaluate different aspects of inter-sentential reasoning ability. Thus jointly evaluating methods on both tasks can potentially lead to a better understanding of the strength and weakness of each method. To this end, we further build a decoder aiming to faithfully reconstruct the original text given its corresponding sentence embedding from the \textsc{BERT} encoder. Presumably, the output vector of semantic matching (i.e., the predicted sentence embedding $\hat h_k$) can be passed to this decoder for generation.
    
    Specifically, we initialize the decoder with a generative language model, \textsc{Gpt-2} \cite{Radford2019LanguageMA}. The sentence embedding from \textsc{Bert} is fed to \textsc{Gpt-2} as the embedding of the zeroth token. Then, we have the decoder generate a sequence of tokens in the hope that the sequence reconstructs the original sentence.
    We fine-tune the parameters of \textsc{Gpt-2} to minimize the negative log-likelihood loss of reconstructing the original sentence. Note that the decoder is separately trained, and thus has no impact on the detection tasks. During training of the sentence-level transformer, we simply perform auxiliary semantic matching in the latent space as previously described, and thus the incoherence detection is not affected. The decoder is applied only at inference time to convert the predicted sentence representations $\hat h_k$ to text.
	
	\paragraph{DAE Fine-tuning} The above generation approach is based on the following desirable properties of the latent sentence embedding: $(i)$ cycle-consistency \cite{zhu2017unpaired}; the original sentence can be recovered from its latent representation. $(ii)$ local smoothness; nearby latent vectors represent sentences with similar semantics. 
	
	We suspect that the features learned by the \textsc{Bert} encoder may lose information required for sentence reconstruction since the masked language model objective does not enforce cycle-consistency, thus hurting the generation. To remedy this problem, we further fine-tune the \textsc{Bert} encoder together with the \textsc{Gpt-2} decoder as an autoencoder aiming to construct a bijective mapping between sentences and their latent semantic representations. To improve the local smoothness of the latent space \cite{li2020optimus,shen2020educating}, we train the models using a denoising autoencoding (DAE) objective \cite{vincent2008extracting} with a similar noising scheme as in \newcite{lewis2019bart} (permutation ratio 20\%, mask ratio 20\%, and random ratio 20\%). The encoder from the fine-tuned DAE can replace the original \textsc{Bert} encoder for mapping sentences into latent representations. Note that the DAE fine-tuning affects the incoherence detection results. 
	
	
	\section{Experiments}
	\label{exp-setup}
	\begin{table*}[t]
		\centering
		\small
		\begin{tabular}{l|c|c|c|c|c|c|c|c|c|c}
			\hline
			\multirow{2}{*}{Model}& \multicolumn{5}{c|}{\textsc{TimeTravel}} & \multicolumn{5}{c}{\textsc{TripAdvisor}} \\
			\cline{2-11}
			& Acc. & P & R & F1 & AUC & Acc. & P & R & F1 & AUC \\
			\hline
			\hline
			\multicolumn{11}{c}{{\em Missing Sentence Detection}} \\
			\hline
		\textit{Human baseline} & 84.6 & 82.3 & 68.8 & 74.8 & - &79.0&81.6&61.3&70.0&-\\
			\hline
			\textit{Token-level} & \textbf{81.2} &\textbf{73.7} &\textbf{68.0} &\textbf{70.7} &86.9 &72.6 &68.6 &58.0 &62.9 &81.4\\
			\hline
\textit{Sentence-level} & 75.5 &67.1 &52.0 &58.6 &80.1 &75.8 &77.6 &55.5 &64.7 &83.1\\
~~~+ SM & 74.6 &65.1 &51.2 &57.3 &80.8 &\textbf{78.0} &\textbf{80.8} &\textbf{59.0} &\textbf{68.2} &\textbf{83.2}\\
~~~+ SM + pre-train & 80.5 &75.0 &62.2 &68.0 &\textbf{87.1} &75.2 &74.1 &58.5 &65.4 &\textbf{83.2}\\
~~~+ DAE & 72.5 &60.8 &49.2 &54.4 &74.0 &72.4 &72.8 &49.5 &58.9 &79.0\\
~~~+ DAE + SM & 72.4 &61.9 &44.7 &52.0 &75.7 &69.6 &68.2 &45.0 &54.2 &77.4\\
~~~+ DAE + SM + pre-train & 79.0 &72.0 &60.5 &65.8 &84.8 & 74.2 &73.8 &55.0 &63.0 &82.2\\
			\hline
			\hline
			\multicolumn{11}{c}{{\em Discordant Sentence Detection}} \\
			\hline
			\textit{Human baseline} & 95.5 & 88.2 & 89.4 & 88.8 & - &83.5&70.0&59.7&64.5&-\\
			\hline
			\textit{Token-level} & \textbf{90.8} &77.3 &\textbf{76.7} &\textbf{77.0} &\textbf{95.1}&74.8 &48.8 &16.8 &25.0 &69.4\\
			\hline
\textit{Sentence-level} & 87.3 &71.2 &61.2 &65.9 &89.0&79.6 &64.6 &\textbf{40.8} &\textbf{50.0} &79.0\\
~~~+ SM & 87.5 &73.5 &59.0 &65.5 &89.3&79.4 &69.6 &31.2 &43.1 &79.0\\
~~~+ SM + pretrain & 89.2 &\textbf{78.1} &64.2 &70.5 &92.0& \textbf{80.4} &\textbf{70.1} &37.6 &49.0 &\textbf{79.8}\\
~~~+ DAE & 86.9 &69.0 &62.2 &65.4 &88.5&75.8 &53.0 &28.0 &36.6 &74.0\\
~~~+ DAE + SM & 87.5 &70.8 &64.2 &67.4 &89.7&77.0 &57.6 &30.4 &39.8 &74.7\\
~~~+ DAE + SM + pretrain & 89.0 &76.5 &65.0 &70.3 &91.7&79.0 &64.7 &35.2 &45.6 &78.6\\
\hline
		\end{tabular}
		\caption{Experimental results on Missing Sentence Detection (upper) and Discordant Sentence Detection (lower). ``+ SM'' indicates joint detection and semantic matching training. ``+ pre-train'' indicates that the model is first pre-trained on the \textsc{Stories} corpus. ``P'' and ``R'' stand for precision and recall scores, respectively.}
		\label{results-msd-dsd}
	\end{table*}
	\subsection{Experimental Setup}
	\paragraph{Implementation Details} Most components of our models are initialized with pre-trained \textsc{Bert} or \textsc{Gpt-2}. They have the same size and configuration as the original \textsc{Bert} or \textsc{Gpt-2} from the HuggingFace Transformers library \cite{Wolf2019HuggingFacesTS} (``bert-base-cased'' and ``gpt2''). The only exception is the sentence-level transformer, which is learned from scratch with random initialization. To make a fair comparison between sentence-level and token-level approaches, the sentence-level transformers have the same architecture as \textsc{Bert}. The MLPs have one hidden layer and the hidden state size is $768$.
	
	To pre-train the sentence-level transformers, we use the \textsc{Stories} dataset \cite{Trinh2018ASM}, which is a subset of the CommonCrawl dataset and contains 400M sentences in total. Most documents in \textsc{Stories} are narratives with long chains of coherent events. We use the \textsc{Stories} dataset for two purposes: $(i)$ to fine-tune the sentence embeddings with the denoising autoencoder objective; $(ii)$ to pre-train our sentence-level models. For the former purpose, we split the documents in \textsc{Stories} into individual sentences. For the latter, we extract text segments of $16$ contiguous sentences from the documents in \textsc{Stories} and use a missing/replacing rate of $25\%$. Other implementation details can be found in the Appendix.
	
	\paragraph{Evaluation Metrics} We view incoherence detection as a series of binary classification problems for individual sentences or sentence boundaries. Following the common practice of reporting classification performance, we provide a set of quantitative evaluation results, including precision, recall, and F1 scores. We also draw Receiver Operating Characteristic (ROC) curves and report the Areas Under the Curves (AUC) \cite{fawcett2006introduction}.
	
	For the side text generation task introduced in \ref{text-gen}, we follow \newcite{galley2019grounded} and perform automatic evaluation using standard reference-based metrics, including BLEU \cite{papineni2002bleu}, NIST \cite{doddington2002automatic}, and METEOR \cite{lavie2007meteor}. We also use Entropy \cite{zhang2018generating} and Dist-$n$ \cite{li-etal-2016-diversity} to evaluate lexical diversity.
	\begin{table*}[t]
\centering
\small
\begin{tabular}{c|c|rr|rr|r|r|rr|r}
\hline
\multirow{2}{*}{Dataset}& \multirow{2}{*}{Sentence-level Method}  & \multicolumn{2}{c|}{NIST}   & \multicolumn{2}{c|}{BLEU(\%)}   & \multicolumn{1}{c|}{METE-} & \multicolumn{1}{c|}{Ent.} & \multicolumn{2}{c|}{Dist(\%)}   & \multicolumn{1}{c}{Len.} \\

 &  & \multicolumn{1}{c}{N-2} & \multicolumn{1}{c|}{N-4} & \multicolumn{1}{c}{B-2} & \multicolumn{1}{c|}{B-4} & \multicolumn{1}{c|}{OR(\%)}  & \multicolumn{1}{c|}{E-4} & \multicolumn{1}{c}{D-1} & \multicolumn{1}{c|}{D-2} & \multicolumn{1}{c}{} \\
 \hline
 \hline
	\multicolumn{11}{c}{{\em Sentence Generation for MSD   }} \\
\hline
  \multirow{3}{*}{\textsc{TimeTravel}} & SM + pre-train & 1.19 & 1.19 & 6.34   & 1.09   & \textbf{9.42}& \textbf{9.87} & 2.94   & \textbf{12.51}  & 14.92 \\
   &  DAE + SM + pre-train  & \textbf{1.43} & \textbf{1.44} & \textbf{7.48}   & \textbf{1.28}   & 8.16& 8.39 & \textbf{3.40}   & 11.92  & 9.26  \\
    &Human reference  & - & -  & - & - & - & 10.44& 11.00  & 43.12  & 9.50  \\
    \hline
   \multirow{3}{*}{\textsc{TripAdvisor}} & SM + pre-train  & 0.96 & 0.97 & 5.65   & 0.89   & \textbf{8.41}& \textbf{8.76} & 0.63   & \textbf{2.86}   & 17.05 \\  
  & DAE + SM + pre-train & \textbf{1.27} & \textbf{1.29 }& \textbf{7.23}   & \textbf{1.24}   & 7.43& 6.85 & \textbf{0.69}   & 2.82   & 10.66 \\
   & Human reference & - & -  & - & - & -  & 12.30& 6.68   & 36.64  & 11.71 \\
  \hline
   \hline
	\multicolumn{11}{c}{{\em Sentence Generation for DSD }} \\
\hline
  \multirow{3}{*}{\textsc{TimeTravel}} & SM + pre-train & 1.31 & 1.33 & 7.98   & 1.91   & 10.45   & \textbf{10.01}& 3.11   & 14.00  & 14.85 \\
    & DAE + SM + pre-train  & \textbf{2.59} & \textbf{2.68} & \textbf{18.48}  & \textbf{6.84}   & \textbf{13.66}   & 9.76 & \textbf{5.06}   & \textbf{20.72}  & 9.59  \\
   & Human reference  & - & - & - & - & - & 10.45& 12.08  & 47.45  & 8.90  \\
    \hline
 \multirow{3}{*}{\textsc{TripAdvisor}} & SM + pre-train    & 1.44 & 1.47 & 9.49   & 2.37   & 11.05   & 10.22& 1.12   & 5.90   & 16.49 \\
  & DAE + SM + pre-train & \textbf{2.71} & \textbf{2.83} & \textbf{19.43}  & \textbf{6.92}   & \textbf{13.69}   & \textbf{10.79}& \textbf{1.91}   & \textbf{11.31}  & 11.89 \\
  & Human reference & - & - & - & - & - & 12.30& 6.66   & 36.41  & 11.66  \\  
\hline
\end{tabular}
\caption{Generation evaluation. ``Ent.'' and ``Len.'' refers to Entropy and the average generation length, respectively. } \vspace{-3mm}
\label{results-gen}
\end{table*}
	\subsection{Results and Discussions}
	\paragraph{Missing Sentence Detection} Table \ref{results-msd-dsd} presents our results on the \textsc{TimeTravel} and \textsc{TripAdvisor} test sets. We observe the following:
	\begin{itemize}[wide=0\parindent,noitemsep,topsep=0em]
	    \item The performance of the token-level baseline is generally better than that of the sentence-level baselines. This is unsurprising as the sentence-level baselines compress sentences into vector representations, resulting in the loss of fine-grained inter-sentence token dependencies that the token-level baseline can capture.
	    \item Joint training of MSD and Semantic Matching (SM) can slightly improve the detection performance, implying that they can work synergistically. This multi-task learning strategy leads to a performance boost as the detection and semantic matching tasks share the same underlying encoder and sentence-level transformer. This indicates that understanding \textit{what is missing} is important to help determine \textit{whether a sentence is missing}.  
	    \item Pre-training on a large corpus significantly improves the performance of the sentence-level baselines, leading to a prediction accuracy comparable to that of the token-level baseline, with much less computational cost at inference time. 
	    \item The original \textsc{Bert} embeddings perform slightly better than the DAE-fine-tuned embeddings. We speculate that the DAE fine-tuning alters the geometry of the latent embedding space for better reconstruction or generation capability while slightly sacrificing the discriminative features for the detection tasks. However, the large-scale pre-training diminishes the difference. With pre-training, the DAE embeddings even outperform the \textsc{Bert} embeddings on the \textsc{TripAdvisor} dataset.
	\end{itemize}
	\paragraph{Discordant Sentence Detection}
	Table \ref{results-msd-dsd} presents the results for DSD, which are largely consistent with our findings in the MSD experiments. Additionally, we observe the following:
	\begin{itemize}[wide=0\parindent,noitemsep,topsep=0em]
	\item Joint training of DSD and semantic matching gives detection performance comparable to that of the detection-only models.
	\item The token-level baseline performs worse than sentence-level baselines on the \textsc{TripAdvisor} dataset. The reason might be that the relatively long input sequences ($8$ sentences) are not handled well by the token-level model.
	\end{itemize}
	\paragraph{Computational Cost} We compare the forward speed of baseline models. For the DSD task of \textsc{TripAdvisor}, the speed of the vanilla sentence-level model is about $860$ paragraphs/s, while the speed of the token-level model is about $40$ paragraphs/s on a single K80 GPU. The computational overhead for SM is negligible. The pre-training takes $\sim20$ hours on $8$ V100 GPUs, but can be reused once trained.
	\paragraph{Sentence Generation Side Task} As the incoherence detection and the corresponding generation tasks associate tightly with each other, to provide a full spectrum of evaluation beyond our main detection task, we further evaluate our sentence-level baselines on the counterpart generation tasks.
	
	We compare two sentence-level models (with or without DAE-fine-tuning), and use beam search with a width of $5$ for generating sentences in the ground-truth missing/discordant positions in our test sets. The results are shown in Table \ref{results-gen}. We have the following observations:
	\begin{itemize}[wide=0\parindent,noitemsep,topsep=0em]
	\item The DAE-fine-tuned embeddings outperform the original \textsc{Bert} embeddings in almost all relevance metrics on both datasets and both tasks. For the diversity scores, the DAE results are comparable to or even better than those from the \textsc{Bert} embeddings.
	\item The performance gap between the DAE-fine-tuned embeddings and original \textsc{Bert} embeddings is more obvious on the generation task corresponding to DSD than that to MSD.
	\item The average length of generation from DAE is closer to the ground truth.
	\end{itemize}
	For qualitative measure of the generation quality, we show some examples in the Appendix.
    \section{Related Work}
	\paragraph{Inter-sentential Reasoning and Understanding}
	Inter-sentential reasoning and understanding have been studied in different forms, including classic discourse parsing tasks using the RST Discourse Treebank \cite{carlson2003building} and the Penn Discourse Treebank \cite{prasad2008penn}. \newcite{chen-etal-2019-evaluation} propose a suite of tasks including sentence position, binary sentence ordering, discourse coherence classification, and sentence section prediction. Narrative cloze tasks \cite{Chambers2008UnsupervisedLO, Mostafazadeh2016ACA} aim to find the right concluding sentence for an incomplete narrative. \newcite{Bhagavatula2020AbductiveCR} formulate abductive commonsense reasoning in order to decide the most plausible hypothesis that could explain the transition between two observations. Our task makes a unique contribution to this area. Importantly, it enjoys the simplicity of data creation yet potentially enables evaluation of various inter-sentential reasoning aspects.

	\paragraph{Contextualized Text Infilling}
	Our task is also related to recent work studying the problem of contextualized text infilling. A number of models \cite{Fedus2018MaskGANBT,song2019mass,liu-etal-2019-tigs,Zhu2019TextI,lewis2019bart,joshi-etal-2020-spanbert,shen2020blank} have been proposed for generating missing text spans in context. However, they focus more on local fluency than inter-sentential semantic coherence. Recently, \newcite{kang-hovy-2019-linguistic} propose the bridging task to generate intermediate sentences between the two given sentences and \newcite{Huang2020INSETSI, wang2019t} explore the task of \textit{sentence infilling}, the task of predicting the intermediate missing sentence that can semantically bridge the surrounding context. However, all of them require the positions of the missing spans to be pre-specified. Some also explore context-aware text modification such as simplification \cite{biran-etal-2011-putting} and style transfer \cite{cheng-etal-2020-contextual,shih2019xl}. Again, though, the input must specify the position to modify. In general, most previous work focuses solely on the problem of \textit{what to generate} and has not considered the problem of \textit{where to generate}. This is insufficient for practical applications such as editing assistance, since we may not know the positions of missing spans or problematic sentences \textit{a priori} \cite{mori2020finding}. Our work fills in the gap between existing work on contextualized text infilling and real-world needs.
	\section{Conclusion}
	We introduced the task of narrative incoherence detection, where the goal is to identify any semantic discrepancy (missing or discordant sentences) in a narrative. Besides its practical value in editing assistance, this task can also be used as a benchmark task for inter-sentential semantic understanding. We hope our work can facilitate future research.
	
\section*{Ethical Impact}
This work focuses on benchmarking models for their capability to capture sentence-level semantic incoherence. The aim of this work is to advance natural language processing (NLP) and general artificial intelligence (AI) research. Our work can be leveraged to evaluate discourse-level natural language understanding (NLU) models and to further shed light on the future development of new models. The corresponding text generation tasks can also encourage the development of natural language generation (NLG), especially semantic planning and reasoning for the long-form text generation. We identify and summarize the ethical considerations including the benefits and potential risks of this work as follows.

This work can facilitate research on the aforementioned NLU/NLG tasks in a generic manner. Such development in NLP can potentially bring improvements in helping models adhere to a reasonable narrative flow and potentially help enforce the generation to obey social norm and fairness, reducing the chance of hallucinating facts, and is thus of the best interests of the general public. 

We also note that this work is a fundamental research work that focuses on model evaluation and technical improvements. Thus, we have NOT applies additional aggressive filtering techniques to the text data we use, beyond what has been performed to the original datasets from their sources. The text data we use may have offensiveness/toxicity/fairness/bias issues that we do not identify, as they are not the main focus of this work. 

Given the aforementioned potential risks and due to the nature of NLG models, we note that the generations or outputs of this work, though not likely, may reflect gender or other historical biases implicit in the data. In rare circumstances, the generations may exhibit a mild extent of unethical, biased, or offensive attitude. These are known issues in current state-of-the-art text generation models. We hope that a better control of the narrative coherence as what we pursue can enable researchers to further investigate these issues and develop mitigation strategies.
	\bibliography{emnlp2021}
	\bibliographystyle{acl_natbib}
\end{document}